\documentclass{article}
\usepackage{spconf,amsmath,graphicx}
\DeclareMathOperator*{\argmax}{arg\,max}
\usepackage{mathrsfs}
\usepackage{caption}

\title{Visual Reranking with improved image graph}
\name{Ziqiong~Liu$^{\star}$ \qquad Shengjin~Wang$^{\star}$ \qquad Liang~Zheng$^{\star}$ \qquad Qi~Tian$^{\dagger}$}

\address{$^{\star}$ Tsinghua University, Beijing, 100084, China \\
    $^{\dagger}$ University of Texas at San Antonio, TX, 78249, USA}

\begin{document}
%
\maketitle
\begin{abstract}
This paper introduces an improved reranking method for the Bag-of-Words (BoW) based image search.
Built on \cite{fusion}, a directed image graph robust to outlier distraction is proposed.
In our approach, the relevance among images is encoded in
the image graph, based on which the initial rank list is refined.
Moreover, we show that the rank-level feature fusion can be adopted in this reranking method as well.
Taking advantage of the complementary nature of various features, the reranking performance is further enhanced. Particularly, we exploit the
reranking method combining the BoW and color information.
Experiments on two benchmark datasets demonstrate that our method yields significant improvements and the reranking results
are competitive to the state-of-the-art methods.

\end{abstract}
\section{Introduction}
\label{sec:intro}
This paper considers the task of the Bag-of-Words (BoW) based image search,
especially on the visual reranking. In BoW model, visual words
are generated using unsupervised clustering algorithms \cite{AKM,HKM} on local features such as
SIFT descriptor \cite{sift,three}.
Then an image is represented as a histogram of visual words.
Basically, each visual word is
weighted using the \textit{tf-idf} scheme \cite{S03,zhenglp}, and the fast search is
achieved through an inverted file.

Nevertheless, traditional BoW model is not satisfying due to many reasons, \textit{e.g.}, the lack of spatial information \cite{knn,he,LZ,SC}, the information loss due to feature quantization \cite{lost,he,zheng2014bayes}, \textit{etc}. Furthermore, BoW model is limited when facing the challenges of occlusions, or viewpoint and illumination changes.

\begin{figure}
    \centering
    \includegraphics[width=0.9\linewidth]{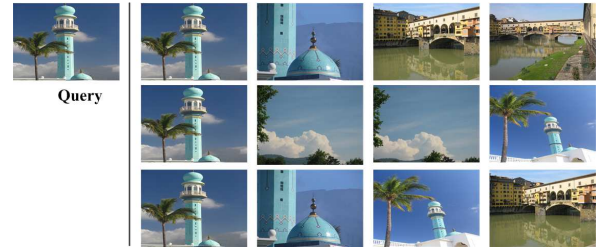}
   \caption{A sample query from the Holidays dataset and its retrieval results
   obtained by BoW (Top), global color histogram (Middle) and proposed reranking (Bottom) combing BoW and color features.}
   \label{fig:1}
\end{figure}

Besides, the BoW based image search aims to find the query's nearest neighbors in SIFT space.
However, SIFT only describes the local texture feature, and the image search system built on a single feature suffers
from low recall, \textit{i.e}., images similar in other feature spaces are not considered, such as color \cite{COMI,color} and semantics \cite{C13, fisher}.
For example, although images containing the same object are distant in SIFT space due to the change of viewpoint, they may be adjacent in color space (see Fig. \ref{fig:1} for an illustration).

In this paper, we tackle these problems using a graph-based method. We construct a directed image graph connecting each image with its potential relevant images. In this way, true matched images, \textit{e.g}., images containing the same object but with different viewpoints, would be linked together. Then, through the graph analysis, the initial results are reranked. What is more, feature fusion at rank-level can be adopted in the graph-based reranking as well. Complementary nature of various features further boosts the performance.

Our work relates to the recent study of visual reranking using image-level cues.
 To name a few, $k$-$NN$ reranking \cite{knn} refines the initial rank list automatically using the $k$-nearest neighbors. Alternatively, Qin \textit{et al.} \cite{hello} take advantage of $k$-reciprocal nearest neighbors to identify the image set for reranking. In addition, a lot of works conduct the reranking based on graph theory \cite{fusion,PAMI,icassp2013,CVPRgraph}, which have shown promising performance.

Due to the limitation of single feature, some works explore the reranking using complementary features in graph-based scheme.
Particularly, Zhang \textit{et al.} \cite{fusion} propose a fusion method  for a specific query to combine the strengths of holistic and local features at rank level. Similarly, Deng \textit{et al.} \cite{ICCV13} introduce a weakly
supervised multi-graph learning framework for visual reranking.

Our work is built on \cite{fusion}, which achieves the state-of-the-art performance by the fusion of
different features using undirected graph.
However, we find \cite{fusion} is sensitive to outlier distraction. Here, outliers mean the irrelevant images of query contained in the graph.
Specifically, when the main parameter $k$, the number of nearest neighbors used in graph construction, is not appropriate, an image node may be connected with many irrelevant images or outliers.
In this situation, \cite{fusion} does not perform well.
In comparison, our improved image graph is more robust to outlier distraction, and yields better reranking performance.

The remainder of the article is organized as follows.
In Section \ref{sec:method}, we describe our reranking
method in details.
Experiments are shown in Section \ref{sec:experiment}.
We conclude in Section \ref{sec:conclusion}.

\section{Our Approach}
\label{sec:method}
In this section, we first introduce the original image graph in Section \ref{subsec:original}. After that we illustrate our algorithms in Section \ref{subsec:construction}, \ref{subsec:fusion} and \ref{subsec:ranking}.
We show that our method aims to select more
candidates to improve recall when constructing graph. Furthermore, the discriminative edge
weight and robust ranking algorithm ensure the precision.

\begin{figure}
\centering
    \includegraphics[width=0.9\linewidth]
    {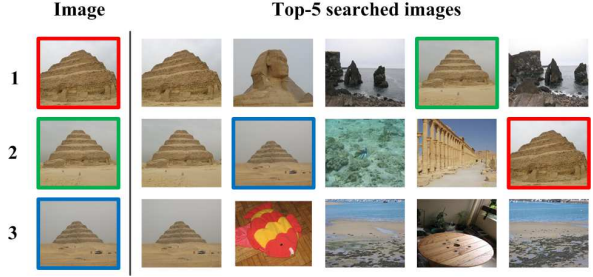}
   \caption{Difference of graph construction between original and improved methods. Image 1 and 2 are relevant and they satisfy \emph{$R_5(1,2)$}. Although Image 2 and 3 are relevant, they do not satisfy \emph{$R_5(2,3)$} for obvious viewpoint change. Thus, the original image graph only contains  image 1 and  2. In comparison, since image 3 is included in $N_5(2)$, the improved image graph would also contain image 3. As a result, the improved method preserves more potential relevant images.}
   \label{fig:2}
\end{figure}

\begin{figure}
\centering
    \includegraphics[width=0.9\linewidth]
    {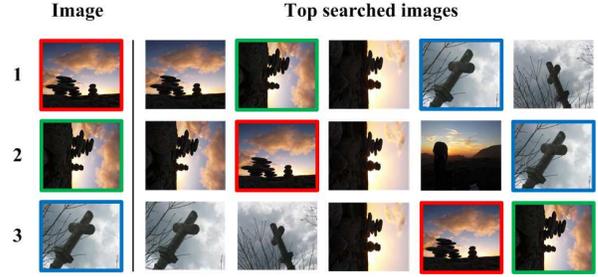}
   \caption{Difference of edge weight between original and improved methods. Let decay coefficient be 1 for observation convenience.
   If $k$ is 3, we can get $w(1,2)$=1 and $w(1,3)$=0 according to Eq. \ref{2}. However, if $k$ is 5, we have $w(1,2)$=2/3 but $w(1,3)$=1. It illustrates edge weight in Eq. \ref{2} is sensitive to $k$. Besides, when $k$ is inappropriate, the edge linking irrelevant images may have larger weight than relevant images. In comparison, we get $w(1,2)$=1/4 no matter $k$ is 3 or 5 according to Eq. \ref{3}. When $k$ is 5, $w(1,3)$ becomes 1/8 smaller than the weight $w(1,2)$ of relevant images.}
   \label{fig:3}
\end{figure}

\subsection{Original image graph}
\label{subsec:original}
 Zhang \textit{et al.} \cite{fusion} propose an undirected image graph for merging global and local features. This method finds potential relevant images based on reciprocal neighbor relation.
 Let $N_{k}(i)$ be the set of $k$ nearest neighbors of image $i$, then the reciprocal neighbor relation is defined as follows:
\begin{equation}\label{1}
  R_k(i,i')=i\in N_{k}(i')\wedge i'\in N_{k}(i)
\end{equation}

The image graph can be represented as $\mathcal{G}$=\{$\mathcal{V}$,$\mathcal{E}$,$w$\},
where $\mathcal{V}$ is the set of images containing query $q$ and $\mathcal{E}$ is the set of edges linking images. There is an undirected edge linking images $i$ and $i'$ if they satisfy Eq. \ref{1}.
The edge weight $w$ is determined by neighborhood consistency of the connected images, and can be written as:

\begin{equation} \label{2}
w(i,i')=
\left\{
\begin{array}{ll}
	\alpha(q,i,i')\frac{|N_{k}(i)\cap N_{k}(i')|}{|N_{k}(i)\cup N_{k}(i')|}& \textrm{if} \ (i,i')\in R_k(i,i')\\
	0 & \textrm{otherwise}\\
\end{array}\right.
\end{equation}
where $|\cdot|$ represents the cardinality of the set and $\alpha(q,i,i')$ is a decay coefficient.
Let $\delta(q,i)$ denote the length of shortest path between $q$ and $i$ in $\mathcal{G}$ , and the decay coefficient can be defined as $\alpha(q,i,i')=\alpha_0^{max(\delta(q,i),\delta(q,i'))}$.
Usually, the $\alpha_0$ is set as 0.8.

For a single graph or a fused graph combining multiple features, ranking is achieved by searching a subgraph $\mathcal{G'}$ with  maximum weighted density .

\subsection{Construction of improved image graph}
\label{subsec:construction}
It is mentioned in \cite{accurate} that reciprocal neighbor relation is a reliable indicator for two images being visually similar.
Therefore, Eq. \ref{1} is a relatively strict  constraint for relevant images, and could filter out many potential irrelevant images during graph construction. However, some relevant images with obvious visual difference may not satisfy Eq. \ref{1}.
Consequently, the graph constructed using reciprocal neighbor relation tends to lose potential candidates.

To further preserve the possible candidates or relevant images when constructing graph, we propose a directed image graph to encode the relevance among images.
In this method, there is a directed edge from image $i$ to image $i'$ if $i'$ appears in the $N_k(i)$.
And the ``directed'' means $i$ may point to $i'$ when $i'$ does not point to $i$.
As a result, more relevant cues among images, \textit{e.g.}, the relevant images filtered out by Eq. \ref{1},  are preserved in the directed graph (see the example in Fig. \ref{fig:2}).

Since the directed graph improves the recall, many outliers are also introduced.
Therefore, it is a critical issue how to define discriminative edge weight.

Edge weight $w$ in Eq. \ref{2} is defined by consistency among top candidates. However, as can be observed in Fig. \ref{fig:3}, the weight heavily depends on $k$.
Besides,
when $k$ is inappropriate, irrelevant images may have more common nearest neighbors compared to relevant ones. It leads to the fact that edge linking irrelevant images obtains larger weight than relevant images. Hence, the discriminative power of the edge weight is limited.

To alleviate the impact of parameter $k$, we propose to define the weight using retrieval ranks.
Let $Rank(i,i')$ be the rank of image $i'$ when using image $i$ as the query. In consideration of
the reciprocal relation, image $i$ and its neighbor $i'$ are close to each other if $Rank(i,i')$ and $Rank(i',i)$ are both high.

Therefore, the edge weight is determined by:

\begin{equation} \label{3}
w(i,i')=
\left\{
\begin{array}{ll}
	\frac{\alpha(q,i,i')}{Rank(i,i')+Rank(i',i)}& \textrm{if} \ i' \in N_k(i)\\
	0 & \textrm{otherwise}\\
\end{array}\right.
\end{equation}

The edge weight defined by Eq. \ref{3} is less dependent on $k$, as it only considers the reciprocal ranks instead of neighborhood consistency. Thus, when $k$ is not chosen properly, the edge weight is still discriminative compared to the original method (see Fig. \ref{fig:3}).


\subsection{Fusion}
\label{subsec:fusion}
Following the basic criterion, it is easy to construct  directed image graphs $\mathcal{G}_n$=\{$\mathcal{V}_n$,$\mathcal{E}_n$,$w_n$\} for different retrieval results. Then we fuse multiple graphs into one graph $\mathcal{G}$=\{$\mathcal{V}$,$\mathcal{E}$,$w$\} without supervision \cite{fusion} , which can be written as:
\begin{equation}\label{4}
  \mathcal{E}=\cup_{n}\mathcal{E}_n,
  \mathcal{V}=\cup_{n}\mathcal{V}_n
\end{equation}

\begin{equation}\label{5}
w(i,i')=\Sigma_{n} w_n(i,i')
\end{equation}

The fusion of various features based on graph can bring many benefits.
First, more candidates are provided to improve the recall by combining the advantages of different features.

Besides, positive images visually similar to query are easier to search, no matter in color or texture space.
Graph fusion prompts these images to link the query with larger weighted edge.
On the other hand, usually negative images cannot be searched in both feature space, hence these images may get smaller edge weight.
In this way, multi-graph fusion insures the precision.

\subsection{Ranking}
\label{subsec:ranking}
For a graph obtained by single feature or multiple features, the relevant
probability of the connected images is encoded into the edge weight. Intuitively, we aim to find the
subgraph $\mathcal{G'}$ containing $q$ in $\mathcal{G}$, which satisfies
the following condition:
\begin{equation}\label{6}
  \mathcal{G'}=\argmax\limits_{\mathcal{G'}=\{\mathcal{V'},\mathcal{E'},w\},q\in \mathcal{V'}}\sum_{(i,i')\in\mathcal{E'}}w(i,i')
\end{equation}
To solve Eq. \ref{6}, we first define the node set $\mathcal{S}$=\{q\} and $\mathcal{C}$ containing nodes $\mathcal{S}$ points to.
The node linked by the largest weighted edge in $\mathcal{C}$ is introduced into $\mathcal{S}$. After that the node sets $\mathcal{S}$ and $\mathcal{C}$ are updated. This procedure continues until cardinality of $\mathcal{S}$ satisfies user's requirement. The nodes are ranked according to their order of insertion into $\mathcal{S}$.

Different from \cite{fusion}, we only consider
maximizing local weighted instead of maximizing weighted density.
As a consequence, the ranking method is less affected by outlier distraction and guarantees the precision of reranking.

\section{Expriments}
\label{sec:experiment}

\subsection{Datasets}
In this paper, we evaluate our proposed method on two public datasets, INRIA Holidays \cite{he} and UKBench \cite{HKM}.
The Holidays dataset consists of 1491 images and 500 of them are queries. Most queries have less than 4 ground truth images undergoing various changes. Retrieval accuracy is measured by mAP (mean average precision).
The UKBench dataset contains 10200 images. Every 4 images are taken from the same object with different viewpoints and illuminations. The N-S score is calculated to measure retrieval accuracy, which refers to the average recall of the top four ranked images.

\subsection{Experiment settings}
 This paper exploits three baselines, which are denoted as BoW, HE and HSV (see Table \ref{table:rerank}).

\textbf{BoW} We adopt approach proposed in \cite{AKM} as BoW baseline. Following \cite{three}, rootSIFT is used on every point. A codebook of size 20K is trained by approximate kmeans \cite{AKM}.

\textbf{HE} We incorporate the weighted Hamming Embedding (HE) \cite{burst} into the baseline of BoW to enhance performance.

\textbf{HSV} We make use of global HSV feature for complementary information. For each image, we compute the 1000 dimension HSV color histogram. Following \cite{COMI}, $L_1$ normalization and square scaling are performed for each color histogram. Retrieval is based on nearest neighbor search using Euclidean distance.

\subsection{Experimental Results }
We apply our approach to three baselines and obtain the reranking results: BoW Graph, HE Graph and HSV Graph, respectively.
The performance of our method on Holidays is illustrated in Fig. \ref{fig:rerank}.
Considering the trade-off between efficacy  and efficiency, we set the number of nearest neighbors \emph{$k$} to 10. After reranking, large improvements over the baseline can be observed from Table \ref{table:rerank}.

Moreover, we fuse HSV cues with BoW and HE separately to further enhance the performance.
The reranking using complementary features improves the baseline significantly (see Table \ref{table:rerank}).
For Holidays, combination of multiple features boosts the baselines of BoW and HE by
26.28\% and 8\% in mAP, respectively.
Similar phenomena are observed on UKBench.
The gain in N-S score over the BoW and HE are 0.755 and 0.311, respectively. Notably, we achieve a mAP of 84.6\% on Holidays and an N-S score of 3.80 on UKBench, which are comparable to the state-of-the-arts (see Table \ref{table:fusion}).

Besides, as we can see in Fig. \ref{fig:rerank}(a), our method is robust to outlier. There are many irrelevant images or outliers in the graph when $k$ is large, since most queries have less than 4 relevant
images. But the accuracy is not affected even if $k$ reaches 60. 
It demonstrates the robustness of this reranking method to outlier distraction.

\begin{table}

\begin{center}
    \begin{tabular}{|c|c|c|}
    \hline
    Methods&Holidays(mAP\%) & UKBench(N-S)\\
    \hline
    BoW&49.16&3.013\\
    HE&76.60&3.491\\
    HSV&63.90&3.398\\
    \hline
    BoW Graph&57.20&3.342\\
    HE Graph&80.97&3.612\\
    HSV Graph&68.16&3.697\\
    \hline
    HSV+BoW &75.44 &3.768  \\
     HSV+HE  &84.60& 3.802 \\
    \hline
    \end{tabular}
\end{center}
    \caption{The performance of reranking.}
    \label{table:rerank}
\end{table}

\begin{table}

\begin{center}
    \begin{tabular}{|c|c c c c c|}
    \hline
    Methods &ours &\cite{fusion} &\cite{burst} &\cite{ICCV13}&\cite{knn}\\
    \hline
    Holidays(mAP\%) & 84.6 & 84.6 &\textbf{84.8} &84.7&- \\
    UKBench (N-S)  & \textbf{3.80} &3.77 &3.64  &3.75&3.52\\
    \hline
    \end{tabular}
\end{center}
    \caption{Comparison with the state-of-the-arts.}
    \label{table:fusion}
\end{table}

\begin{figure}
\centering
    \includegraphics[width=8.6cm]{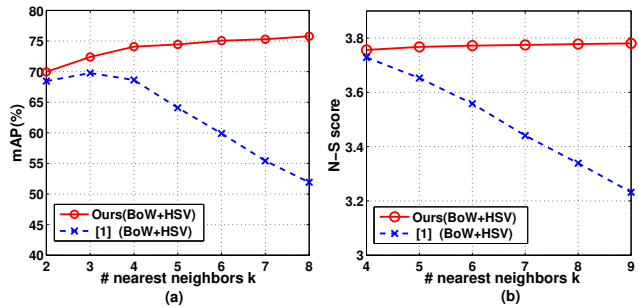}
   \caption{The performance of reranking on (a) Holidays and (b) UKBench when fusing BoW and HSV. }
  \label{fig:fusion}
\end{figure}

A comparison with \cite{fusion} is presented in Fig. \ref{fig:fusion}. When the parameter \emph{$k$} is chosen properly, \cite{fusion} can achieve satisfying performance. However, \cite{fusion} is very sensitive to parameter \emph{$k$} or the outlier distraction. When \emph{$k$} becomes large, the performance of \cite{fusion} decreases. In comparison, our method is robust to outlier distraction and yields better fusion results.

\subsection{Complexity}
\begin{figure}
\centering
    \includegraphics[width=8.6cm]{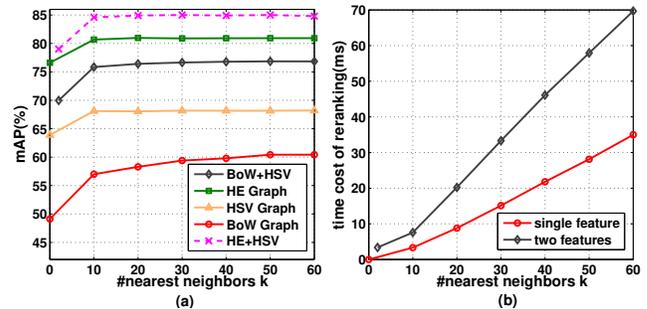}
   \caption{The reranking evaluation of (a) mAP and (b) time cost on Holidays. }
  \label{fig:rerank}
\end{figure}

For graph construction, each image is used as query in our search system. Then, we compute and store their relevant relationships. The memory complexity is $O(Nk)$, where $N$ is the database size.
The running time depends on parameter $k$, and Fig.\ref{fig:rerank}(b) shows the time cost of this reranking using Matlab on a server with 3.46 GHz CPU and 64 GB memory.

\section{Conclusion}
\label{sec:conclusion}
In this paper, we propose an improved image graph for visual reranking, which is
robust to outlier distraction. The graph encodes the relevance among images and the initial rank list
is refined based on the graph. Moreover, this reranking could
adopt multiple features to further enhance the performance.
We have achieved an mAP of 84.6\% on Holidays and an N-S score of 3.80 on UKBench using reranking combining BoW and color information.
Our future work involves experimenting our method on more datasets and exploring more features such as attribute and GIST.

\textbf{Acknowledgements} This work was supported by the National High Technology Research and Development Program of China (863 program) under Grant Nos. 2012AA011004 and the National Natural Science Foundation of China under Grant Nos. 61071135 and the National Science and Technology Support Program under Grant No. 2013BAK02B04.
 This work was also supported in part to Dr. Qi Tian by ARO grant W911NF-12-1-0057,  Faculty Research Awards by NEC Laboratories of America, and 2012 UTSA START-R  Research Award respectively. This work was supported in part by National Science Foundation of China (NSFC) 61128007.

\bibliographystyle{IEEEbib}
\bibliography{strings,refs}

\begin{thebibliography}{10}

\bibitem{fusion}
S.~Zhang, M.~Yang, T.~Cour, K.~Yu, and D.N. Metaxas,
\newblock ``Query specific fusion for image retrieval,''
\newblock in {\em ECCV}, 2012, pp. 660--673.

\bibitem{AKM}
J.~Philbin, O.~Chum, M.~Isard, J.~Sivic, and A.~Zisserman,
\newblock ``Object retrieval with large vocabularies and fast spatial
  matching,''
\newblock in {\em CVPR}, 2007, pp. 1--8.

\bibitem{HKM}
D.~Nister and H.~Stewenius,
\newblock ``Scalable recognition with a vocabulary tree,''
\newblock in {\em CVPR}, 2006, pp. 2161--2168.

\bibitem{sift}
D.G. Lowe,
\newblock ``Distinctive image features from scale-invariant keypoints,''
\newblock {\em IJCV}, vol. 60, no. 2, pp. 91--110, 2004.

\bibitem{three}
R.~Arandjelovic and A.~Zisserman,
\newblock ``Three things everyone should know to improve object retrieval,''
\newblock in {\em CVPR}, 2012, pp. 2911--2918.

\bibitem{S03}
J.~Sivic and A.~Zisserman,
\newblock ``Video {G}oogle: a text retrieval approach to object matching in
  videos,''
\newblock in {\em ICCV}, 2003, pp. 1470--1477.

\bibitem{zhenglp}
L.~Zheng, S.~Wang, Z.~Liu, and Q.~Tian,
\newblock ``Lp-norm {Idf} for large scale image search,''
\newblock in {\em CVPR}, 2013, pp. 1626--1633.

\bibitem{knn}
X.~Shen, Z.~Lin, J.~Brandt, S.~Avidan, and Y.~Wu,
\newblock ``Object retrieval and localization with spatially-constrained
  similarity measure and k-{NN} re-ranking,''
\newblock in {\em CVPR}, 2012, pp. 3013--3020.

\bibitem{he}
H.~J{\'e}gou, M.~Douze, and C.~Schmid,
\newblock ``Hamming embedding and weak geometric consistency for large scale
  image search,''
\newblock in {\em ECCV}, 2008, pp. 304--317.

\bibitem{LZ}
L.~Zheng and S.~Wang,
\newblock ``Visual phraselet: Refining spatial constraints for large scale
  image search,''
\newblock {\em Signal Processing Letters,IEEE}, vol. 20, no. 4, pp. 391--394,
  2013.

\bibitem{SC}
W.~Zhou, Y.~Lu, H.~Li, Y.~Song, and Q.~Tian,
\newblock ``Spatial coding for large scale partial-duplicate web image
  search,''
\newblock in {\em ACM MM}, 2010, pp. 511--520.

\bibitem{lost}
J.~Philbin, O.~Chum, M.~Isard, J.~Sivic, and A.~Zisserman,
\newblock ``Improving particular object retrieval in large scale image
  databases,''
\newblock in {\em CVPR}, 2008, pp. 1--8.

\bibitem{zheng2014bayes}
L.~Zheng, S.~Wang, W.~Zhou, and Q.~Tian,
\newblock ``Bayes merging of multiple vocabularies for scalable image
  retrieval,''
\newblock in {\em CVPR}, 2014.

\bibitem{COMI}
L.~Zheng, S.~Wang, Z.~Liu, and Q.~Tian,
\newblock ``Packing and {P}adding: {C}oupled {M}ulti-index for {A}ccurate
  {I}mage {R}etrieval,''
\newblock in {\em CVPR}, 2014.

\bibitem{color}
C.~Wengert, M.~Douze, and H.~J{\'e}gou,
\newblock ``Bag-of-colors for improved image search,''
\newblock in {\em ACM MM}, 2011, pp. 1437--1440.

\bibitem{C13}
S.~Zhang, M.~Yang, X.~Wang, Y.~Lin, and Q.~Tian,
\newblock ``Semantic-aware {C}o-indexing for {N}ear-duplicate {I}mage
  {R}etrieval,''
\newblock in {\em ICCV}, 2013.

\bibitem{fisher}
M.~Douze, A.~Ramisa, and C.~Schmid,
\newblock ``Combining attributes and {F}isher vectors for efficient image
  retrieval,''
\newblock in {\em CVPR}, 2011, pp. 745--752.

\bibitem{hello}
D.~Qin, S.~Gammeter, L.~Bossard, T.~Quack, and L.~Van~Gool,
\newblock ``Hello neighbor: accurate object retrieval with k-reciprocal nearest
  neighbors,''
\newblock in {\em CVPR}, 2011, pp. 777--784.

\bibitem{PAMI}
Y.~Jing and S.~Baluja,
\newblock ``Visualrank: Applying pagerank to large-scale image search,''
\newblock {\em PAMI}, vol. 30, no. 11, pp. 1877--1890, 2008.

\bibitem{icassp2013}
C.~Huang, Y.~Dong, H.~Bai, L.~Wang, N.~Zhao, S.~Cen, and J.~Zhao,
\newblock ``An efficient graph-based visual reranking,''
\newblock in {\em ICASSP}, 2013, pp. 1671--1675.

\bibitem{CVPRgraph}
W.~Liu, Y.~G. Jiang, J.~Luo, and S.~F. Chang,
\newblock ``Noise resistant graph ranking for improved web image search,''
\newblock in {\em CVPR}, 2011, pp. 849--856.

\bibitem{ICCV13}
C.~Deng, R.~Ji, W.~Liu, D.~Tao, and X.~Gao,
\newblock ``Visual reranking throughweakly supervised multi-graph learning,''
\newblock in {\em ICCV}, 2013.

\bibitem{accurate}
H.~J{\'e}gou, C.~Schmid, H.~Harzallah, and J.~Verbeek,
\newblock ``Accurate image search using the contextual dissimilarity measure,''
\newblock {\em PAMI}, vol. 32, no. 1, pp. 2--11, 2010.

\bibitem{burst}
H.~J{\'e}gou, M.~Douze, and C.~Schmid,
\newblock ``On the burstiness of visual elements,''
\newblock in {\em CVPR}, 2009, pp. 1169--1176.

\end{thebibliography}

\end{document}